# Enhanced Robot Motion Block of A-star Algorithm for Robotic Path Planning

**Raihan Kabir[1], Yutaka Watanobe[2] (Member, IEEE), Md. Rashedul Islam[3] (Senior Member, IEEE) and Keitaro Naruse[4] (Member, IEEE)**

[1,2,4] School of Computer Science and Engineering, University of Aizu, Aizu-wakamatsu, Fukushima, 965-8580, Japan
[3] Chief Researcher of Computer Vision and AI, Chowagiken Corp., Japan

Corresponding author: Yutaka Watanobe, Raihan Kabir (e-mail: yutaka@u-aizu.ac.jp, raihan.kabir.cse@gmail.com).

This paragraph of the first footnote will contain support information, including sponsor and financial support acknowledgment.

**ABSTRACT** An efficient robot path-planning model is vulnerable to the number of search nodes, path cost, and time complexity. The conventional A-star (A*) algorithm outperforms other grid-based algorithms for its heuristic search. However it shows suboptimal performance for the time, space, and number of search nodes, depending on the robot motion block (RMB). To address this challenge, this study proposes an optimal RMB for the A* path-planning algorithm to enhance the performance, where the robot movement costs are calculated by the proposed adaptive cost function. Also, a selection process is proposed to select the optimal RMB size. In this proposed model, grid-based maps are used, where the robot's next move is determined based on the adaptive cost function by searching among surrounding octet neighborhood grid cells. The cumulative value from the output data arrays is used to determine the optimal motion block size, which is formulated based on parameters. The proposed RMB significantly affects the searching time complexity and number of search nodes of the A* algorithm while maintaining almost the same path cost to find the goal position by avoiding obstacles. For the experiment, a benchmarked online dataset is used and prepared three different dimensional maps. The proposed approach is validated using approximately 7000 different grid maps with various dimensions and obstacle environments. The proposed model with an optimal RMB demonstrated a remarkable improvement of 93.98% in the number of search cells and 98.94% in time complexity compared to the conventional A* algorithm. Path cost for the proposed model remained largely comparable to other state-of-the-art algorithms. Also, the proposed model outperforms other state-of-the-art algorithms.

**INDEX TERMS** A* algorithm, Adaptive cost function, BFS, Dijkstra, DFS, Path planning, Robot motion block (RMB).

## I. INTRODUCTION

In the field of robotics, there are several challenging tasks, and path planning is one of the most important tasks among them to find a way to obtain the desired goal position. In this research area, the main objective is to develop an efficient path-planning algorithm by enhancing its performance in terms of searching time complexity, path cost, and search area [1]. As the application area of robot usage is increasing rapidly, the number of ongoing studies is also increasing in this area [2]. Generally, path-planning algorithms always search for the target position from the current position of the robot by keeping the overall path as shortest as possible. There are many path-planning algorithms for minimizing the path cost and finding the shortest path; however, for a path-planning algorithm, the importance of the three main factors, time complexity, path cost, and search area, mainly depends on the robot application [1]. Nevertheless, considering these factors, a general and efficient path-planning algorithm fits most robot applications.

Generally, path-planning algorithms are categorized into two groups based on the surroundings of the known degree of environmental information global path planning and local path planning of the robot [3, 4]. A robot with global path planning needs to know the environmental information partially, such as a map of the environment required to calculate the optimal route from the start position to the goal position. The algorithm partially knew the obstacle map and the effectiveness of these algorithms is evaluated by their ability to find a valid path if one exists [5, 6]. In contrast, local path planning requires the robot to search its surroundings using sensors and find a path to reach the goal in an environment with unknown information [5]. As a result, local path-planning algorithms are typically more time-consuming and complex and may require the use of techniques such as segmentation and object detection to avoid obstacles [7]. Some commonly used local path-planning algorithms include the artificial potential field method, fuzzy logic method, and dynamic window method. In some cases, an ArUco markers-based searching system can make the process easier and more efficient [8]. In the global path-planning field, various algorithms are available, such as rapidly exploring random trees (RRTs), graph search, tangent graph-based planning, and predefined path optimization. Our research focus is on the area of graph-based searching algorithms.

Graph-based searching algorithms are extensively used and considered highly effective in the field of global path planning. Some popular examples of graph search algorithms include A*, Dijkstra, depth-first search (DFS), and breadth-first search (BFS). Among these algorithms, A* is particularly notable for its completeness, efficient search area, and ability to use heuristics [9]. A* is a path-finding algorithm developed on the basis of the Dijkstra algorithm and is designed for use on weighted graphs, such as grid maps with obstacles. The algorithm works by constructing a tree of paths from the starting node and adding nodes as it searches the graph until it reaches the goal node or exhausts all options [9]. The final result is the shortest path found in the tree of searched paths. To search the graph or grid map, the algorithm requires a RMB to navigate the surrounding environment. The research goal is to develop an optimal RMB to improve the overall performance of the A* algorithm.

The robot motion block is a crucial element in path-planning algorithms such as A*, as it determines how the robot will search the cells surrounding its current position as it moves toward its destination [10]. The RMB is a matrix that specifies the cells that the robot should search in each iteration to find the goal node starting from the start node. The performance and efficiency of the algorithm are significantly affected by the size of the motion block matrix, as it determines the search area of the robot in each iteration [11]. An inefficient motion block results in the robot searching for a larger number of cells to find the goal, increasing the time and space complexity. In the conventional approach, the motion block matrix comprises eight surrounding cell coordinates for the robot to search [10]. In each iteration, the A* algorithm adds the information about these eight cells to its search queue and continues to add cells proportionally until it reaches the goal or exhausts all options. This approach necessitates a significant amount of searching time complexity and many search nodes. Thus, keeping these limitations in mind, this study proposes an optimal RMB with an adaptive cost function for robot movement to improve the performance. The proposed method considers eight neighboring coordinates with different searching distances and puts only the end node in the search queue, significantly reducing the number of search nodes and time complexity with a small path cost increment. As the path cost increases and the number of search nodes and time complexity decrease with different sizes of motion block, a formula is also proposed to decide the optimal one among different distances. Using the proposed method, the A* algorithm can find the goal position by avoiding obstacles with significantly improved overall performance. A comparison study with the conventional A* and different similar state-of-the-art path-planning algorithms validates the significant improvement of the proposed model. In addition, this study discusses the effectiveness, drawbacks, and applicability of the proposed approach in different application areas.

The remainder of this paper is structured as follows: Section 2 provides an overview of related works, and Section 3 presents the details of the proposed model. Section 4 offers the discussion and results of the experiments conducted, and Section 5 concludes the study.

## II. RELATED WORKS

In the field of robotics, path-planning algorithms have been extensively investigated by researchers to improve their efficiency and reduce search costs. Despite considerable progress in this area, the performance of these algorithms still has room for improvement. Researchers are still exploring new ways to enhance their performance to make them more reliable and efficient. This is particularly important in robotics applications, where fast and accurate path planning is essential for the safe and effective operation of robots that can handle increasingly complex environments and tasks while maintaining a high-performance level.

In one study, Li Changgeng et al. proposed a new global path-planning algorithm called bidirectional alternating search A* (BAS-A*) for mobile robots [12]. Their algorithm combines the best features of traditional A* and bidirectional search algorithms to generate efficient and optimal paths. By utilizing a bidirectional alternating search strategy, the algorithm improves search efficiency by searching forward and backward path lists until they intersect. The algorithm addresses issues such as long calculation times, large turning angles, and unsmoothed paths in large task spaces. They introduced weighted heuristic functions, a filtering function for path nodes, and Bézier curves to ensure smooth and optimized path planning [13]. Simulation results demonstrated the algorithm's efficiency and smoothness compared to other algorithms. Practical validation was also attempted on the TurtleBot3 Waffle Pi mobile robot [14].

Meanwhile, Szczepanski et al. presented a hybrid approach for the global path planning of mobile robots in variable workspaces [15]. Their approach comprised an offline global path optimization algorithm using the artificial bee colony algorithm and an online path-planning scheme that uses a graph created from the control path points generated in the offline part [16]. The online part uses the Dijkstra algorithm to find the shortest path [17]. Their proposed approach aimed to generate optimal paths for mobile robots in variable workspaces feasible even with new obstacles. Their approach was tested in the Matlab/Simulink environment and achieved a good result.

In another study, Sánchez–Ibáñez et al. conducted a comprehensive review article that provided an overview of path-planning algorithms for autonomous mobile robots [18]. They highlighted the importance of autonomous capabilities in enhancing economic and safety aspects. With numerous path-planning algorithms available, the authors aimed to classify and analyze these algorithms, particularly for autonomous ground vehicles [19]. Their review covered environmental representation, robot mobility, dynamics, and various path-planning categories. The study served as a valuable resource for understanding the research conducted on path-planning algorithms, aiding in the selection of appropriate algorithms for specific requirements.

Meanwhile, Tripathy et al. proposed a collision-free navigation scheme for mobile robots in a grid environment [20]. This research scheme combines a radio frequency identification method for robot localization, a hybrid approach for path planning, and a predefined decision table for navigation [21, 22]. The algorithm has two stages: construction of the virtual world and generation of the optimal shortest path. The performance of the algorithm was evaluated in different grid-based environments with and without obstacles [23]. The results showed that the robot explored fewer cells to find the shortest path when there were no obstacles. However, in environments with obstacles, the number of turns in the shortest path was always less than that when exploring the entire virtual world.

Separately, Chen et al. introduced a three-neighbor search A* algorithm combined with the artificial potential field method to overcome irregular forward direction obstacles and guide robot movement [24, 25]. Their method demonstrated significant improvements by reducing the number of search nodes, search time, and path length by 88.85%, 77.05%, and 5.58%, respectively. Saeed et al. presented the boundary node method, an offline path-planning approach that generated collision-free paths for mobile robots [26]. Their method utilized a nine-node quadrilateral element and a potential function to guide robot movement, generating initial collision-free paths quickly and safely. They also employed a path enhancement method to further optimize the path length. The simulation results validated the effectiveness of their proposed methods. In a different approach, Ichter et al. introduced latent sampling-based motion planning, combining recent control advances with techniques from sampling-based motion planning (SBMP) [27, 28]. Their methodology involved learning a plannable latent representation using autoencoding, dynamics, and collision-checking networks. The learned latent RRT algorithm demonstrated global exploration capabilities and generalization to new environments, as showcased in visual planning and humanoid robot path-planning problems [29]. These studies provide valuable insights into different path-planning approaches that helped us to address the limitations of traditional A* algorithm and offer improved performance and adaptability for various robotic systems.

## III. PROPOSED MODEL

The A* algorithm is most effective in finding the shortest path due to its heuristic search technique. The conventional A* algorithm takes a long searching time and checks a large number of cells for searching the goal because of an inefficient RMB. For a robot to reach its destination, it needs to find the shortest path from the starting position to the goal by checking a minimum number of search nodes and small time complexity. The proposed RMB helps to significantly reduce those two criteria and improves the overall performance of the A* path-planning algorithm while avoiding obstacles. The block diagram in Fig. 1 illustrates the proposed model, which includes a robot path-planning algorithm with an efficient RMB to search for the optimal path. In this model, the robot examines the neighboring grid cells using the proposed optimal RMB, which significantly improves the performance of the path-planning algorithm and increases its usability in real time. The proposed model was experimented using various grid maps that contain different obstacle environments, starting and goal positions.

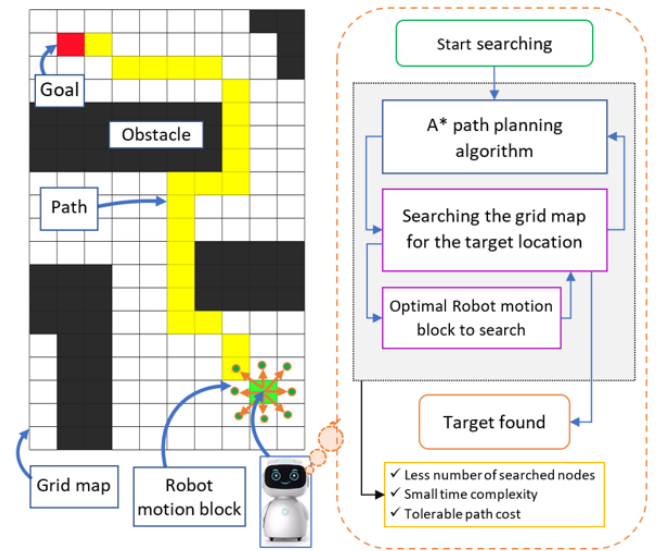

**FIGURE 1.** Block diagram of our proposed model.

## A. A* PATH PLANNING ALGORITHM WITH PROPOSED RMB

Currently, there exist numerous algorithms for searching paths and traversing graphs, one of which is A*. It is extensively recognized for its efficiency in being both optimal and complete. A* is classified as a best-first search or informed search algorithm that operates on a weighted graph. Its objective is to find the goal node, beginning from a specified starting node with the smallest amount of travel cost and time. The algorithm creates a tree of search nodes, beginning from the starting node, and continues until the search terminates, which can either be the discovery of the goal node or no discovery at all. The A* algorithm uses a heuristic function $h(x)$ to estimate the cost $c(s, s')$ to reach the goal node from the current node [30], where $h(x) \leq c(x, x') + h(x')$, and $c(x, x')$ is the cost of an individual node $x \neq x_{goal}$ and any successor $x'$. This heuristic should be both admissible (it should never overestimate the actual cost to reach the goal) and consistent (the cost from one node to another should not be more than the cost to move directly from the first node to the goal node). The heuristic function for the grid map is the Euclidean distance between the current node and the goal node. The algorithmic workflow of the A* algorithm, using the proposed model, shown in Fig. 2, is as follows:

1) To begin the A* algorithm with the proposed robot motion block, initialize the start node ($S_{start}$) with a cost of 0, and add it to an $OPEN$ list of nodes to be considered for expansion.

2) Check the neighboring grid nodes for the goal node ($S_{goal}$) with the proposed robot motion block explained in subsection 3.2. If the current node is an obstacle node ($S_{obstacle}$), ignore it, and add only the end nodes in the $OPEN$ list. While the $OPEN$ list is not empty, select the node with the lowest total cost $Cost(C)$ and remove it from the $OPEN$ list.

$$Cost(C) = min\{OPEN, \; h(S_{\text{goal}}, OPEN)\}$$

Here, $h(S_{\text{goal}}, OPEN)$ denotes a heuristic function that uses Euclidean distance, $d(p, q) = \sqrt{(q_1 - p_1)^2 + (q_2 - p_2)^2}$

3) If the selected node is the goal node, stop the search and return the optimal path. Generate all possible successors of the selected node and calculate their costs and heuristic values.

4) For each successor node, update its cost and heuristic value if it has not already been visited or has a higher cost. Add each updated successor node to the $OPEN$ list if it is not already in the list.

5) If a successor node is already in the $OPEN$ list, update its cost and heuristic value if the new values are lower than the previous values. If a successor node is already in the visited list (i.e., has already been expanded), ignore it.

6) If all nodes in the motion block cells have been checked, return to step II and repeat the steps until every node on the grid map has been checked.

7) Terminate the algorithm when the goal node is found or the $OPEN$ list is empty.

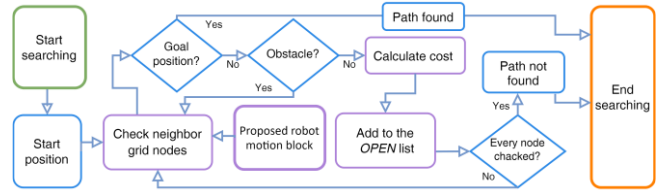

**FIGURE 2.** Block diagram of A* path planning algorithm with proposed optimal RMB.

## B. PROPOSED ROBOT MOTION BLOCK WITH AN ADAPTIVE COST-FUNCTION

In a grid-based search, robots must search the cells around them to find the goal node in the smallest number of cells possible. The robot uses a block of cells, called robot motion block or motion kernel, to search for its neighboring nodes. However, the conventional RMB used by A* leads to a higher number of search cells and a longer search time. The proposed motion block, which surrounds the robot with eight neighboring cells, aims to minimize the number of search cells and the time taken to find the goal position while maintaining a tolerable path cost.

$$Motion\ block = \begin{bmatrix} n & 0 & CC(d,C) \\ 0 & n & CC(d,C) \\ -n & 0 & CC(d,C) \\ 0 & -n & CC(d,C) \\ -n & -n & DC(d',C) \\ -n & n & DC(d',C) \\ n & -n & DC(d',C) \\ n & n & DC(d',C) \end{bmatrix} \quad (1)$$

Here, $\boldsymbol{n} = 1, 2, 3, \ldots, N$

$\boldsymbol{CC}$ = Cardinal movement cost.
$\boldsymbol{DC}$ = Diagonal movement cost.
$\boldsymbol{d}$ = The cost associated with cardinal movements.
$\boldsymbol{d'}$ = The cost associated with diagonal movements.
$\boldsymbol{C}$ = Adaptive cost.

The proposed RMB is represented by a matrix as Equation (1), where $n$ is the size of the RMB, which indicates the distance between the robot and the cells being searched, and $N$ denotes a natural number. The first column represents the x-axis, the second column represents the y-axis, and the third column represents the robot movement cost of each point in a two-dimensional grid space. When $n$ increases, the proposed method checks the nodes till the $n^{th}$ node but only adds the $n^{th}$ node to the $OPEN$ list. Fig. 3 provides a visual representation of the proposed RMBs.

VOLUME XX, 2017

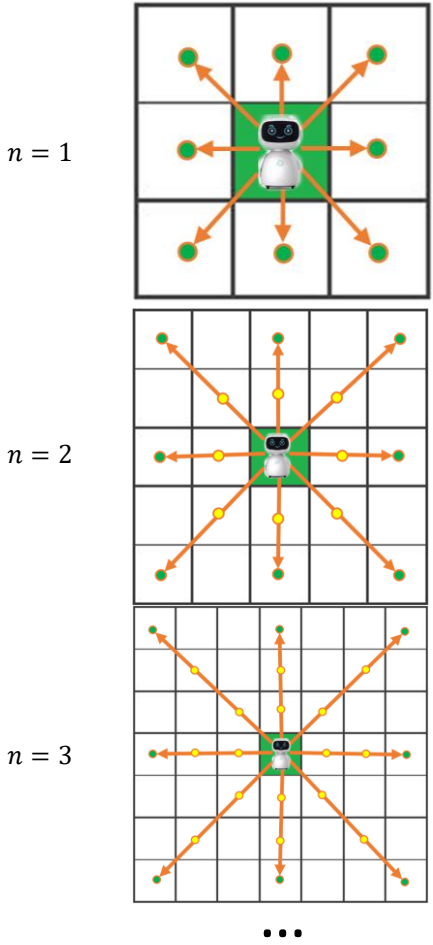

**FIGURE 3.** Block diagram of A* path planning algorithm with proposed optimal RMB.

The RMB has eight neighboring cell directions, including up, down, left, right, up-right, up-left, down-right, and down-left. The costs of the four cardinal and diagonal directed points are represented by $CC(d, C)$, and $DC(d', C)$ respectively. These costs are calculated by the following Equations (2), (3), and (4):

$$d_i = 1, \quad 1 \leq i \leq 4 \text{ and } d'_i = \sqrt[2]{2}, \quad 5 \leq i \leq 8$$

$$C'(Ccn_j, cn_j, q_i) = Ccn_j + \sqrt[2]{(qx_i - cnx_j)^2 + (qy_i - cny_j)^2} \quad (2)$$

Subject to, $1 \leq j \leq N; 1 \leq i \leq 8$; $cn$ = Current node, $Ccn$ = Cost of $cn$, $q$ = Nodes of motion matrix.

$$C(C', gn_i, q_i) = C' + \left(\sqrt[2]{(gnx_i - qx_i)^2 + (gny_i - qy_i)^2}\right) * \alpha \quad (3)$$

Subject to, $gn$ = Goal node, and $0.001 \leq \alpha \leq 0.009$

$$CC(d, C) = d * C,$$
$$DC(d', C) = d' * C, \quad (4)$$

For achieving the costs $CC$ and $DC$ of the nodes of the motion matrix, an adaptive cost function $C$ is formulated, which is multiplied by the costs associated with cardinal $d$ and diagonal $d'$ movements as Equation (4). In this adaptive cost function, first, the current node's cost $Ccn$ is summed with the Euclidian distance between the current node $cn$ and the node of motion matrix $q$ to get the absolute differences in the x and y coordinates. The cost $C'$ is achieved as in Equation (2). Next, $C'$ summed with the Euclidean distance between the node of motion matrix $q$ and goal node $gn$ by multiplying with a constant of $\alpha$ in Equation (3). Here, $\alpha$ is used to increase the importance of the distance value between each of the motion matrix nodes and the goal node. This helps the adaptive cost function to keep track of the goal node with each node of the motion matrix and reduce the number of searches to find the goal node. The optimal range of $\alpha$ is between 0.001 and 0.009 after testing on a large number of maps. However, $\alpha = 0.007$ is used in this paper which provided the best results.

### C. OPTIMAL ROBOT MOTION BLOCK SELECTION

The proposed robot motion block has different sizes ($n$). The output results for each motion block differ. However, a general optimal RMB is essential for any robotic application. To obtain the optimal one from different RMB sizes, a selection formula is proposed as Equations (5), (6) and (7).

$$Ra(A_i)_j = R \times \frac{A(x_j) - (A)_{min}}{(A)_{max} - (A)_{min}} \quad (5)$$

$$Ra'(Ra)_j = \frac{\sum_{i=1}^{p} Ra(x_j)}{p} \quad (6)$$

Subject to, $1 \leq i \leq p, 1 \leq j \leq n, R = 1000, p = 3$

$$ORMB(Ra') = argmin\left(\frac{\sum_{k=1}^{q} Ra'(x_j)}{q}\right) \quad (7)$$

Subject to, $1 \leq k \leq q, \ q = 5$ (map types)

Here, $A$ denotes the array of $n$ elements, each element contains the mean of output data. Which is experimented on the dataset containing $\{1, ..., N\}$ data points, based on the RMB size. $i$ denotes paramiters and $A(x_j)$ denotes individual output array eliments for each RMB size; where $\{x_j, j = 1, ..., n\}$. To optimize the efficiency of the path-planning algorithm, this research considered three paramiters $p$, those are path costs, the number of search cells in the grid map, and time complexity. Next, Each data array is brought into a range of $R = 1000$ using Equation (5) and achieves a ranged array $Ra$. After that, each indexed data from all resultant data arrays is summed and subscribed by the total number of arrays, which in this research is three, to obtain $Ra'$. Equation (6) was used to calculate $Ra'$. Finally, the optimal RMB is determined by selecting the minimum value of the final resultant data array for each RMB. Equation (7) was used to select the optimal RMB $ORMB$.

### IV. EXPERIMENTAL RESULTS AND DISCUSSION

The experimental outcomes justify the validity and demonstrate the effectiveness of the proposed model. The following subsections describe the experimental results of the proposed optimal RMB for the A* algorithm. First, the preprocessing steps of the dataset with different types of grid maps are described. After that, the experimental results using the prepared dataset show the effectiveness of the proposed model.

## A. GRID MAP PREPARATIONS USING THE DATASET
### 1) DATASET
To validate the proposed approach, a collection of grid maps is essential for the experiment on the proposed RMB of the A* path-planning algorithm. This experiment uses a benchmarked online public repository dataset named "motion_planning_datasets" [31]. There are eight types of grid maps in this dataset. Each grid map type contains thousands of PNG format images with a resolution of 201 x 201. Five types of grid maps of planning environments are used among these eight types: alternating_gaps, forest, bugtraq_forest, gaps_and_forest, and mazes. For validation purposes, 4000 grid maps data are used in this research, with 800 grid maps data for each type. Fig. 4 shows some sample images of these grid maps from the dataset.

information. To make the proposed method's efficiency more concrete and generalize the method, the proposed method needs to test on different dimensional grid maps, but the dataset images have a diminution of 201 x 201. To overcome these limitations and fit the dataset with the experiment, the dataset images are preprocessed and prepared the grid maps for the experimental format, as depicted in Fig. 5.

First, three different dimensional types of grid map images are prepared 261 x 261, 462 x 261, and 462 x 462. To prepare a 462 x 261 dimensional map, two same types of dataset map images are added side by side, and for a 462 x 462 dimensional map, four same types of dataset map images are added togather. In this way, 200 maps are achieved for 402 x 402, 400 for 402 x 201, and 800 for 201 x 201, as 800 map images are considered from the dataset.

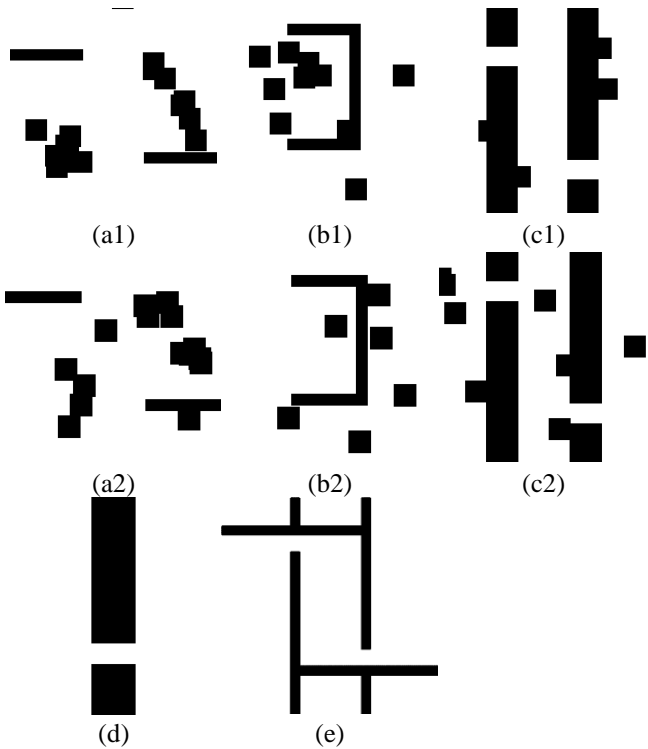

**FIGURE 4.** Sample images of grid maps from the dataset (a1,2) forest,(b1,2) bugtrap_forest, (c1,2) gaps_and_forest, (d) alternating_gaps, and (e) mazes.

### 1) GRID-MAP PREPARATION
The dataset described in the previous subsection has some limitations, such as all grid map data points being images in PNG format rather than the matrix format required for the experiment. For the robot not to leave the grid map, a border surrounding the grid map is essential, but there is no border in the dataset images, and it also has no start and endpoint

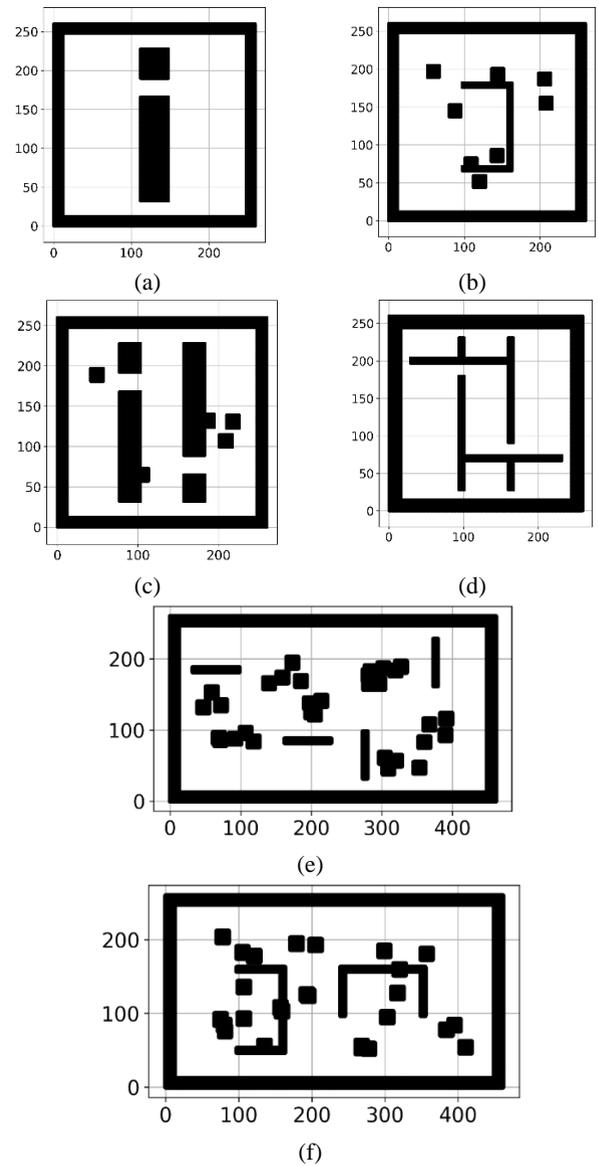

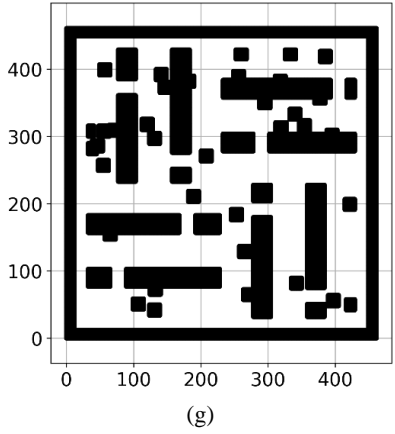

**FIGURE 5.** Prepared grid maps from the dataset (261 x261): (a) alternating_gaps, (b) forest, (c) gaps_and_forest, and (d) mazes; (261 x 462): (e) forest, (f) bugtrap_forest; (462 x 462): (g) gaps_and_forest. Where, white = free cell and black = obstacle cell.

Next, two borders surrounding the dataset grid map images with 15 pixels each are added, making the grid map image dimension 261 x 261, 462 x 261, and 462 x 462. Among these two borders, the outer border is black obstacle pixels, and the inner border is white obstacle-free pixels. Next, these modified map images are split into grid cells containing information about each cell with and without obstacles to make it a matrix format. Finally, the grid maps with the required format are achieved to validate the proposed model.

### 2) SOURCE AND DESTINATION SELECTION

For the experiment, four sets of different sources and destinations have been selected for each of the three-dimensional maps, as shown in Fig. 6. The red arrow represents the direction from the source (depicted as the green circle) to the destination (represented by the blue circle). In addition besides the blue and green circles, source and destination coordinates are provided. For the map with dimensions 261 x 261, 200 maps are experimented with for each direction, as 800 maps are prepared for this dimensional map. Similarly, 100 maps are used for each direction in the case of the 462 x 261 dimensional map, and 50 maps are employed for the 462 x 462 dimensional maps.

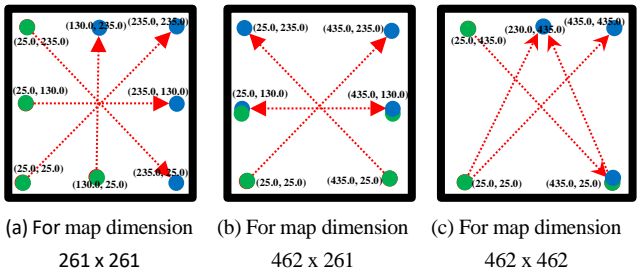

(a) For map dimension 261 x 261  (b) For map dimension 462 x 261  (c) For map dimension 462 x 462

**FIGURE 6.** (a), (b), (c) shows the source to destination direction for three dimension maps'. Where, blue circle = destination, green circle = source, and red arrow = path direction.

### B. EXPERIMENTAL RESULTS OF THE PROPOSED MODEL

This paper shows the experimental results using the preprocessed 4000 grid maps of five different types for the proposed optimal RMB of the A* algorithm. First, the experiment for the proposed model demonstrates the effectiveness of the proposed RMB by minimizing the number of search cells and time complexity while maintaining the minimum path cost from the start position to the target position. Then, the selection of the optimal RMB ensures algorithmic optimality by considering the three major criteria: the number of search cells, time complexity, and path cost. Finally, a comparison between the proposed model and other similar algorithms shows the effective performance of the proposed optimal RMB.

#### 1) COMPARISON BETWEEN CONVENTIONAL A* AND THE PROPOSED METHOD

The proposed method performs better than the conventional A* algorithm because of the adaptive cost function of the robot movement costs, which keeps track of the goal position. When the search distance of the motion block $n$ increases, the performance also increases even more. This is because, in that case, only the end node of this search distance is placed in the next search list. In Table I, the comparison shows the improvement for RMB $n = 1$. However, this paper shows further experiments for other values of RMB.

TABLE I
EXPERIMENTAL RESULTS FOR THE COMPARISON BETWEEN CONVENTIONAL A* AND THE PROPOSED METHOD WITH RMB $n = 1$

| Algorithms | # Search cells | Path cost | Time required (s) |
|---|---|---|---|
| **Conventional A*** | 46878.88 | 390.10 | 13.2112 |
| **Proposed model** (with RMB $n = 1$) | 31604.34 | 389.94 | 9.3736 |

Table I shows the cumulative experimental results conducted using the dataset containing five types of grid maps. Where the proposed model's time complexity reduces by 29.05 %, the number of search nodes reduces by 32.58 %, and path cost reduces by 0.04 % compared to the conventional A* algorithm, calculated using Equation (8).

$$IE = \frac{\frac{\sum_{j=1}^{m} x_j}{m} - \frac{\sum_{j=1}^{m} y_j}{m}}{\frac{\sum_{j=1}^{m} x_j}{m}} \times 100, \qquad (8)$$

Subject to, $\quad 1 \leq j \leq m, m = 3$
(as three dimentional maps are considered)

Here, $IE =$ Impact Evaluation
$x =$ Value of conventional A*
$y =$ Value of proposed model

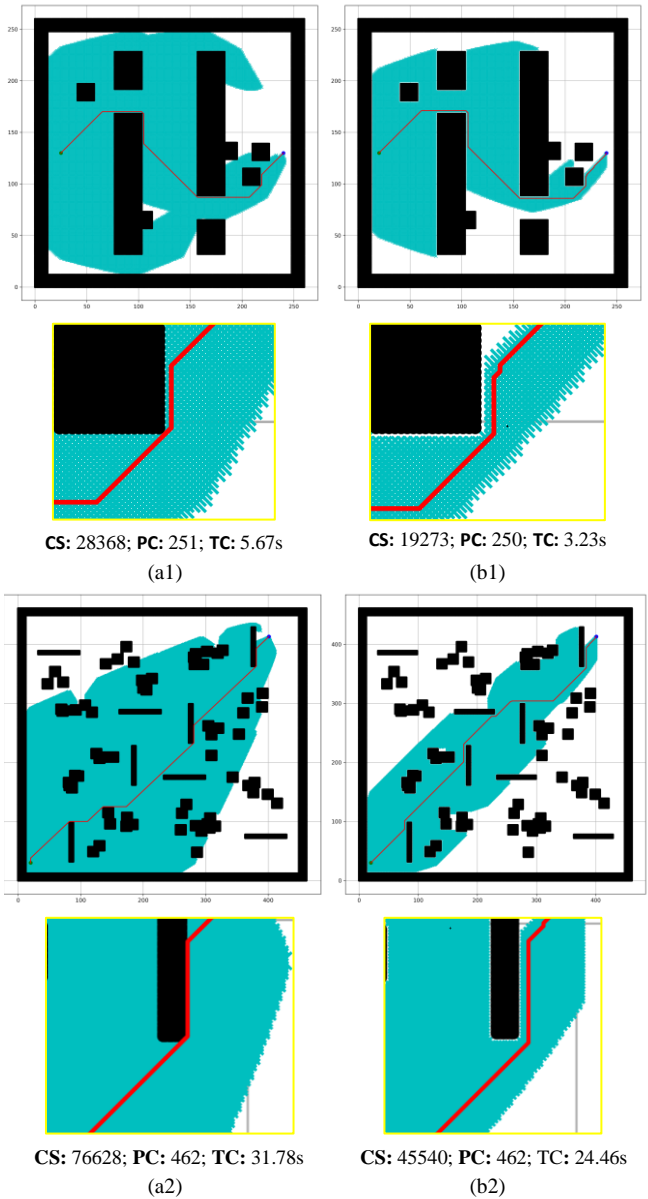

three of the five map types are shown in Fig. 8. (a1–6), (b1–6), and (c1–6) show the result of the bugtraq_forest, gaps_and_forest, and mazes grid maps. Where (a1–6) has a map dimension of 261 x 261, (b1–6) has 462 x 261, and (c1–6) has 462 x 462. To find the target point, the number of search cells drastically decreases with an increase in the size of the proposed RMB. The results for the RMB size $n = 1$ are shown in (a1, b1, c1), those for $n = 2$ in (a2, b2, c2), those for $n = 3$ in (a3, b3, c3), and so on. However, the path cost increased by a tolerable amount.

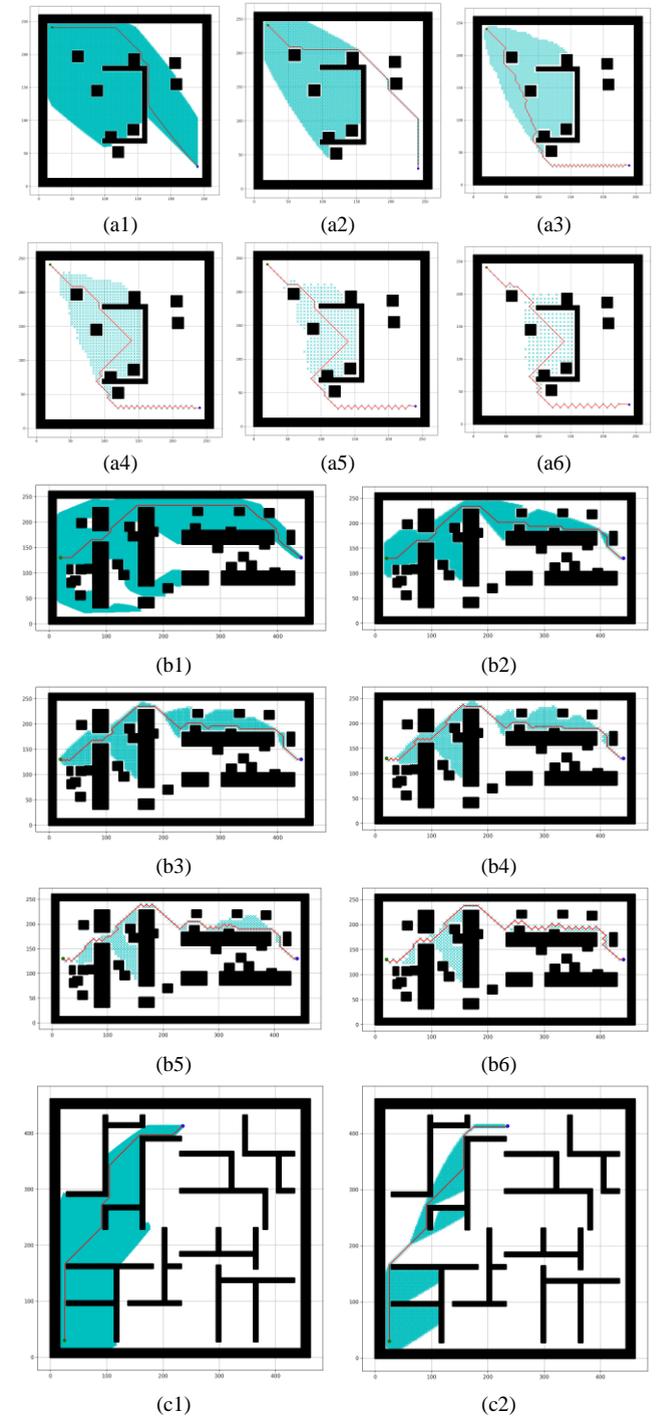

**CS:** 28368; **PC:** 251; **TC:** 5.67s (a1)
**CS:** 19273; **PC:** 250; **TC:** 3.23s (b1)
**CS:** 76628; **PC:** 462; **TC:** 31.78s (a2)
**CS:** 45540; **PC:** 462; **TC:** 24.46s (b2)

**FIGURE 7.** Simulated results for the comparison between (a1, a2) conventional A* and (b1, b2) the proposed method. Here, CS = number of cells searched, PC = path cost, TC = time complexity.

Figure 7 shows improved performance and differences of the proposed model compared to the conventional A* algorithm. It indicates that the proposed model maintains a safe distance from the obstacles where the conventional algorithm goes through close to the obstacles. This is because of its consideration of an offset during the final path calculation.

### 2) EXPERIMENTAL RESULTS USING DIFFERENT SIZES (n) OF THE ROBOT MOTION BLOCK

The simulated outcomes using the proposed RMBs with different sizes ($n = 1$–$6$) for the A* path-planning algorithm are depicted in Fig. 8. The RMB size is assumed ($n = 1$–$6$) for the experiment because if $n$ increases more, it provides mostly similar results as $n = 6$, which does not affect the results. The experimental results for three types of map dimensions and

VOLUME XX, 2017

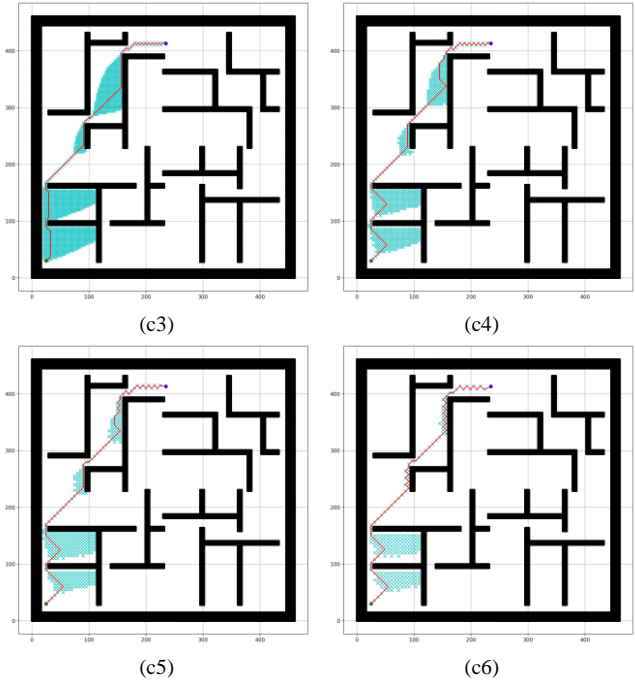

**FIGURE 8.** Simulated experimental outcomes of the proposed RMBs (n = 1–6) for three types of maps. Here, white = free cell, black = obstacle cell, green & blue circle = start & goal point, red line = found path, and cyan cross = cell searched.

In Fig. 8, the red line indicates the path that the proposed approach found toward the target node. Furthermore, the searched grid cells are marked by a cyan cross, of which the number decreases when $n$ increases.

TABLE II
EXPERIMENTAL RESULTS FOR THE A* PATH-PLANNING ALGORITHM USING THE PROPOSED RMB

| Map type | Map dimension | Parameters | RMB | | | | | |
|---|---|---|---|---|---|---|---|---|
| | | | n=1 | n=2 | n=3 | n=4 | n=5 | n=6 |
| alternating_gaps | 261 X 261 | # search cells | 11903.92 | 2838.90 | 1198.65 | 903.76 | 786.34 | 502.14 |
| | | Path cost | 246.26 | 245.78 | 246.01 | 247.32 | 262.92 | 261.95 |
| | | Time required (s) | 2.5259 | 0.2006 | 0.0735 | 0.0585 | 0.0296 | 0.0166 |
| | 462 X 462 | # search cells | 38132.17 | 8351.43 | 2789.37 | 1702.49 | 1076.50 | 888.37 |
| | | Path cost | 431.22 | 433.96 | 435.06 | 436.84 | 443.55 | 445.29 |
| | | Time required (s) | 11.0979 | 0.8327 | 0.1473 | 0.0712 | 0.0383 | 0.0221 |
| forest | 261 X 261 | # search cells | 8181.34 | 1809.13 | 1142.62 | 869.21 | 629.31 | 482.98 |
| | | Path cost | 235.69 | 241.27 | 242.17 | 245.92 | 244.84 | 248.83 |
| | | Time required (s) | 1.9034 | 0.0982 | 0.0521 | 0.0256 | 0.0245 | 0.0195 |
| | 462 X 462 | # search cells | 22056.06 | 3855.53 | 1764.34 | 1356.10 | 1175.00 | 936.63 |
| | | Path cost | 417.87 | 419.61 | 419.07 | 419.09 | 420.10 | 422.81 |
| | | Time required (s) | 5.8014 | 0.4410 | 0.1067 | 0.0617 | 0.0336 | 0.0200 |
| | 462 X 462 | # search cells | 41497.96 | 1098.45 | 759.14 | 632.76 | 567.44 | 495.21 |
| | | Path cost | 423.38 | 425.47 | 425.23 | 427.45 | 429.58 | 428.90 |
| | | Time required (s) | 19.1736 | 0.1115 | 0.0405 | 0.0245 | 0.0186 | 0.0156 |
| bugtraq_forest | 261 X 261 | # search cells | 18436.83 | 5909.82 | 2731.87 | 1533.71 | 1105.00 | 892.15 |
| | | Path cost | 260.13 | 261.72 | 260.52 | 264.59 | 268.00 | 268.97 |
| | | Time required (s) | 3.9576 | 0.4435 | 0.1137 | 0.0500 | 0.0383 | 0.0245 |
| | 462 X 462 | # search cells | 41725.85 | 9255.84 | 4571.77 | 2855.64 | 2372.93 | 1382.53 |
| | | Path cost | 436.70 | 440.48 | 440.36 | 442.83 | 446.93 | 449.65 |
| | | Time required (s) | 12.5779 | 0.8432 | 0.2342 | 0.1087 | 0.0692 | 0.0383 |
| | 462 X 462 | # search cells | 46582.38 | 5011.74 | 1913.54 | 1509.35 | 1339.76 | 1121.14 |
| | | Path cost | 460.24 | 465.91 | 467.55 | 474.02 | 479.76 | 481.96 |
| | | Time required (s) | 15.7666 | 0.8953 | 0.1778 | 0.0861 | 0.0588 | 0.0291 |
| gaps_and_forest | 261 X 261 | # search cells | 9260.30 | 1858.17 | 938.19 | 761.81 | 558.93 | 380.48 |
| | | Path cost | 251.04 | 250.87 | 253.21 | 255.33 | 259.90 | 263.07 |
| | | Time required (s) | 2.3573 | 0.1129 | 0.0455 | 0.0334 | 0.0206 | 0.0186 |
| | 462 X 462 | # search cells | 31361.01 | 7935.98 | 3511.17 | 2356.03 | 1336.57 | 991.89 |
| | | Path cost | 448.15 | 449.54 | 453.95 | 462.83 | 469.79 | 470.84 |
| | | Time required (s) | 6.6211 | 0.6473 | 0.1781 | 0.0847 | 0.0411 | 0.0245 |
| | 462 X 462 | # search cells | 41334.98 | 5584.34 | 2427.80 | 1517.65 | 1124.17 | 759.52 |
| | | Path cost | 485.89 | 486.88 | 488.99 | 493.71 | 497.64 | 496.01 |
| | | Time required (s) | 11.3063 | 0.9169 | 0.1356 | 0.0595 | 0.0393 | 0.0208 |
| mazes | 261 X 261 | # search cells | 26045.38 | 10176.25 | 5986.73 | 3896.61 | 2564.64 | 2008.82 |
| | | Path cost | 320.51 | 320.78 | 322.35 | 325.45 | 329.83 | 330.33 |
| | | Time required (s) | 4.7198 | 0.6722 | 0.2536 | 0.1198 | 0.0658 | 0.0435 |
| | 462 X 462 | # search cells | 49235.53 | 8426.30 | 3654.47 | 2122.42 | 1590.48 | 1247.94 |
| | | Path cost | 471.47 | 472.21 | 473.32 | 475.49 | 481.91 | 500.58 |
| | | Time required (s) | 13.6747 | 0.7576 | 0.1655 | 0.0679 | 0.0479 | 0.0346 |
| | 462 X 462 | # search cells | 58241.65 | 17939.49 | 7454.32 | 4472.24 | 3274.23 | 2194.61 |
| | | Path cost | 522.04 | 524.76 | 527.02 | 530.56 | 532.45 | 536.13 |
| | | Time required (s) | 17.9080 | 1.8732 | 0.3839 | 0.1729 | 0.1012 | 0.0563 |

For the proposed RMB $n = (1–6)$, three types of data were collected from the experimental results: path cost, number of search cells, and the time required (seconds) to find the path. Table II shows the resultant data for each RMB size and map dimension, which is a mean of all grid maps for each type of map among five types. In addition, a comparison of the outcome results is shown in Fig. 9, 10, and 11 where 9(a), 10(a), 11(a) show the number of nodes searched, 9(b), 10(b), 11(b) shows the path cost, and 9(c), 10(c), 11(c) shows the time required to find the target.

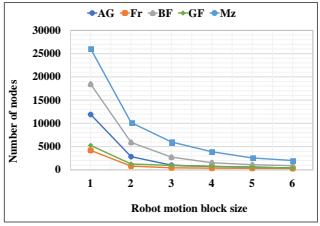

(a) Number of search nodes.

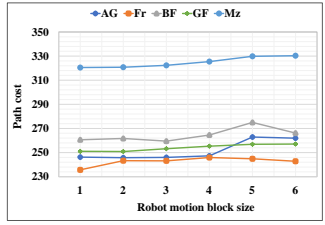

(b) Path cost.

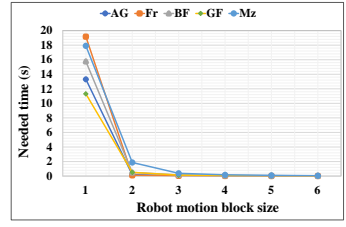

(c) Time required for searching the goal node.

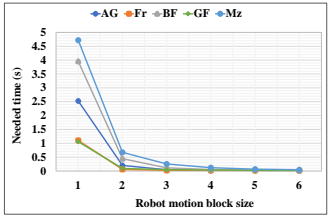

(c) Time required for searching the goal node.

**FIGURE 9. Results for the proposed RMBs for different types of maps where map dimension was 261 x 261. Here, AG = alternating_gaps, Fr = forest, BF = bugtrap_forest, GF = gaps_and_forest, and Mz = mazes.**

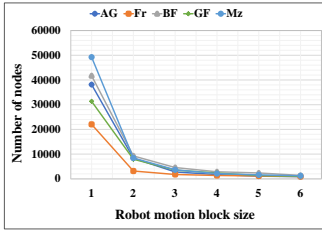

(a) Number of search nodes.

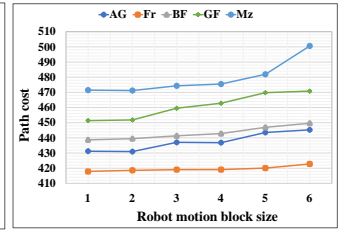

(b) Path cost.

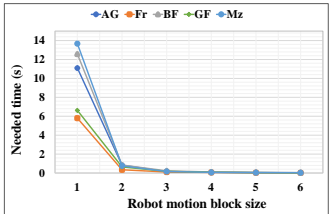

(c) Time required for searching the goal node.

**FIGURE 10. Results for the proposed RMBs for different types of maps where map dimension was 462 x 261. Here, AG = alternating_gaps, Fr = forest, BF = bugtrap_forest, GF = gaps_and_forest, and Mz = mazes.**

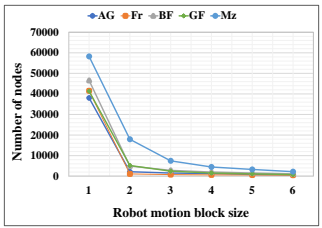

(a) Number of search nodes.

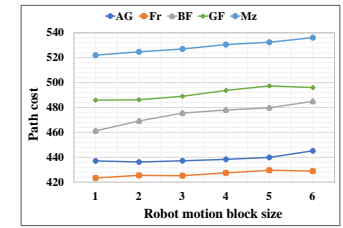

(b) Path cost.

**FIGURE 11. Results for the proposed RMBs for different types of maps where map dimension was 462 x 462. Here, AG = alternating_gaps, Fr = forest, BF = bugtrap_forest, GF = gaps_and_forest, and Mz = mazes.**

The results in Table II and Fig. 9, 10, and 11 show that, with the increase of distance for the RMB, the time complexity and number of search nodes to find the target decreased, whereas the path cost increased. Therefore, an optimal RMB selection is essential and is performed using the proposed formula in Equations (5), (6), and (7), as described in subsection III (C).

TABLE III
EXPERIMENTAL RESULTS TO SELECT THE OPTIMAL RMB USING EQUATIONS (5), (6), AND (7).

| Map type | Map dimension | Parameters | RMB | | | | | |
|---|---|---|---|---|---|---|---|---|
| | | | $n=1$ | $n=2$ | $n=3$ | $n=4$ | $n=5$ | $n=6$ |
| alternating_gaps | 261 X 261 | # search cells | 1000 | 204.95 | 61.09 | 35.22 | 24.93 | 0.00 |
| | | Path cost | 28.30 | 0.00 | 13.42 | 89.74 | 1000 | 943.69 |
| | | Time required (s) | 1000 | 73.30 | 22.66 | 16.67 | 5.16 | 0.00 |
| | | *Average* | **676.10** | **92.75** | **32.39** | **47.21** | **343.36** | **314.56** |
| | 462 X 261 | # search cells | 1000 | 200.38 | 51.04 | 21.86 | 5.05 | 0.00 |
| | | Path cost | 0.00 | 194.97 | 273.07 | 399.64 | 876.12 | 1000 |
| | | Time required (s) | 1000 | 73.18 | 11.31 | 4.44 | 1.46 | 0.00 |
| | | *Average* | **666.67** | **156.18** | **111.80** | **141.98** | **294.21** | **333.33** |
| | 462 X 462 | # search cells | 1000 | 88.85 | 23.79 | 16.96 | 4.78 | 0.00 |
| | | Path cost | 66.33 | 0.00 | 79.17 | 208.90 | 389.30 | 1000 |
| | | Time required (s) | 1000 | 57.59 | 6.82 | 1.36 | 0.32 | 0.00 |
| | | *Average* | **688.78** | **48.81** | **36.59** | **75.74** | **131.47** | **333.33** |
| forest | 261 X 261 | # search cells | 1000 | 172.27 | 85.69 | 50.17 | 19.01 | 0.00 |
| | | Path cost | 0.00 | 424.75 | 493.11 | 779.07 | 696.73 | 1000 |
| | | Time required (s) | 1000 | 41.75 | 17.29 | 3.24 | 2.65 | 0.00 |
| | | *Average* | **666.67** | **212.92** | **198.70** | **277.49** | **239.46** | **333.33** |
| | 462 X 261 | # search cells | 1000 | 138.21 | 39.19 | 19.86 | 11.29 | 0.00 |
| | | Path cost | 0.00 | 351.94 | 241.79 | 246.73 | 450.17 | 1000 |
| | | Time required (s) | 1000 | 72.83 | 15.00 | 7.21 | 2.35 | 0.00 |
| | | *Average* | **666.67** | **187.66** | **98.66** | **91.27** | **154.61** | **333.33** |
| | 462 X | # search cells | 1000 | 14.71 | 6.44 | 3.35 | 1.76 | 0.00 |
| | | Path cost | 0.00 | 337.47 | 298.24 | 656.74 | 1000 | 890.39 |

| | | | 1 | 2 | 3 | 4 | 5 | 6 |
|---|---|---|---|---|---|---|---|---|
| | 462 | Time required (s) | 1000 | 5.00 | 1.30 | 0.46 | 0.15 | 0.00 |
| | | *Average* | **666.67** | **119.06** | **101.99** | **220.18** | **333.97** | **296.80** |
| bugtraq_forest | 261 X 261 | # search cells | 666.67 | 212.92 | 198.70 | 277.49 | 239.46 | 333.33 |
| | | Path cost | 666.67 | 212.92 | 198.70 | 277.49 | 239.46 | 333.33 |
| | | Time required (s) | 666.67 | 212.92 | 198.70 | 277.49 | 239.46 | 333.33 |
| | | *Average* | **666.67** | **190.91** | **57.38** | **182.62** | **302.01** | **333.33** |
| | 462 X 261 | # search cells | 1000 | 195.16 | 79.05 | 36.51 | 24.55 | 0.00 |
| | | Path cost | 0.00 | 291.88 | 282.17 | 472.96 | 789.82 | 1000 |
| | | Time required (s) | 1000 | 64.19 | 15.63 | 5.61 | 2.46 | 0.00 |
| | | *Average* | **666.67** | **183.74** | **125.62** | **171.70** | **272.28** | **333.33** |
| | 462 X 462 | # search cells | 1000 | 85.58 | 17.43 | 8.54 | 4.81 | 0.00 |
| | | Path cost | 0.00 | 260.96 | 336.57 | 634.61 | 898.72 | 1000 |
| | | Time required (s) | 1000 | 55.04 | 9.45 | 3.63 | 1.89 | 0.00 |
| | | *Average* | **666.67** | **133.86** | **121.15** | **215.59** | **301.80** | **333.33** |
| gaps_and_forest | 261 X 261 | # search cells | 1000 | 166.41 | 62.81 | 42.94 | 20.10 | 0.00 |
| | | Path cost | 14.27 | 0.00 | 191.82 | 365.81 | 740.76 | 1000 |
| | | Time required (s) | 1000 | 40.32 | 11.52 | 6.36 | 0.85 | 0.00 |
| | | *Average* | **671.42** | **68.91** | **88.72** | **138.37** | **253.90** | **333.33** |
| | 462 X 261 | # search cells | 1000 | 228.66 | 82.96 | 44.92 | 11.35 | 0.00 |
| | | Path cost | 0.00 | 60.99 | 255.46 | 647.02 | 953.69 | 1000 |
| | | Time required (s) | 1000 | 94.41 | 23.30 | 9.13 | 2.53 | 0.00 |
| | | *Average* | **666.67** | **128.02** | **120.57** | **233.69** | **322.52** | **333.33** |
| | 462 X 462 | # search cells | 1000 | 118.91 | 41.12 | 18.68 | 8.99 | 0.00 |
| | | Path cost | 0.00 | 83.76 | 264.05 | 665.66 | 1000 | 860.81 |
| | | Time required (s) | 1000 | 79.40 | 10.17 | 3.43 | 1.64 | 0.00 |
| | | *Average* | **666.67** | **94.02** | **105.11** | **229.26** | **336.87** | **286.94** |
| mazes | 261 X 261 | # search cells | 1000 | 339.79 | 165.49 | 78.54 | 23.12 | 0.00 |
| | | Path cost | 0.00 | 27.82 | 187.67 | 503.39 | 949.12 | 1000 |
| | | Time required (s) | 1000 | 134.45 | 44.93 | 16.32 | 4.78 | 0.00 |
| | | *Average* | **666.67** | **167.35** | **132.70** | **199.42** | **325.67** | **333.33** |
| | 462 X 261 | # search cells | 1000 | 149.59 | 50.15 | 18.22 | 7.14 | 0.00 |
| | | Path cost | 0.00 | 25.29 | 63.31 | 138.00 | 358.44 | 1000 |
| | | Time required (s) | 1000 | 53.01 | 9.60 | 2.44 | 0.97 | 0.00 |
| | | *Average* | **666.67** | **75.96** | **41.02** | **52.89** | **122.18** | **333.33** |
| | 462 X 462 | # search cells | 1000 | 280.92 | 93.84 | 40.64 | 19.26 | 0.00 |
| | | Path cost | 0.00 | 192.96 | 353.21 | 604.99 | 739.05 | 1000 |
| | | Time required (s) | 1000 | 101.78 | 18.35 | 6.53 | 2.52 | 0.00 |
| | | *Average* | **666.67** | **191.89** | **155.14** | **217.39** | **253.61** | **333.33** |

3) RESULTS FOR OPTIMAL ROBOT MOTION BLOCK SELECTION

Table III presents the range data for each criterion: the number of search cells, path cost, and time complexity. The mean of each map type and map dimension shows that in most cases, the size of RMB $n = 3$ has the minimum numeric value, and it provides the optimal performance of the A* algorithm. Fig. 12 also shows the result for the proposed optimal RMB (which is $n = 3$). This is because it has the lowest value for both the number of search nodes and time complexity while still having a low value for the path cost. Therefore, it balances the three criteria well and provides an optimal overall performance. It also indicates that for the map type, *gaps_and_forest* $n = 2$ is optimal because of its complex and dense obstacle environments. But the differences between the value of $n = 2$ and $n = 3$ is not much, such as for the map dimension 462 x 462 the values are 94.02 and 105.11 respectively.

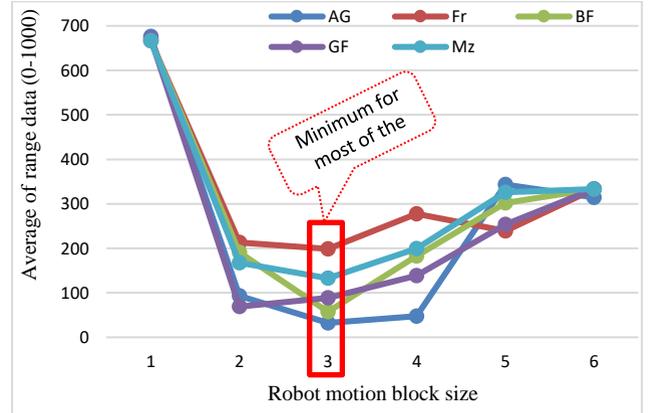

(a) For map dimension 261 x 261.

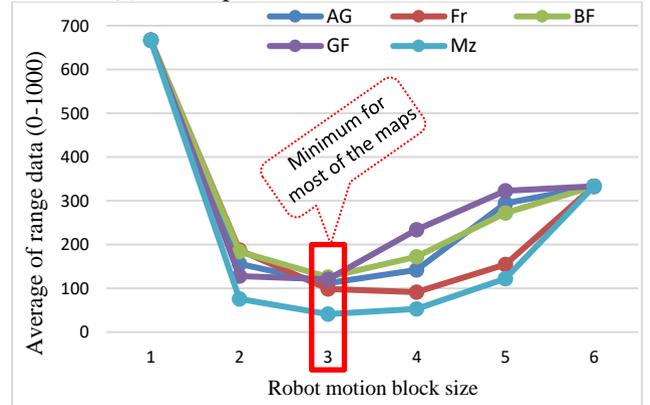

(b) For map dimension 462 x 261.

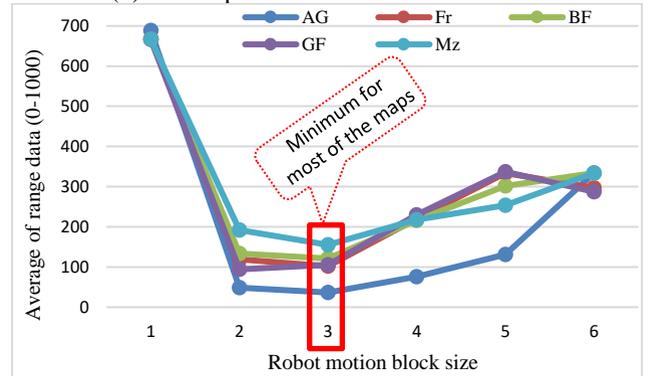

(c) For map dimension 462 x 462.

**FIGURE 12.** (a), (b) and (c) are the experimental results for the proposed optimal RMB based on the experimental data of Table 3.

### 4) COMPARISON OF THE PROPOSED METHOD WITH STATE-OF-THE-ART ALGORITHMS

Figure 13 shows the resultant figures for different state-of-the-art algorithms on a grid map presented in Fig. 13(a). In this grid map, the total number of grid cells was 212521, the number of free cells was 142483, and the number of obstacle cells was 70038. The overall performance of Dijkstra and BFS was mostly similar, but the performance of DFS was the worst in terms of path cost. The performance of the conventional A* algorithm was good, but the proposed model with optimal RMB outperforms all comparison algorithms.

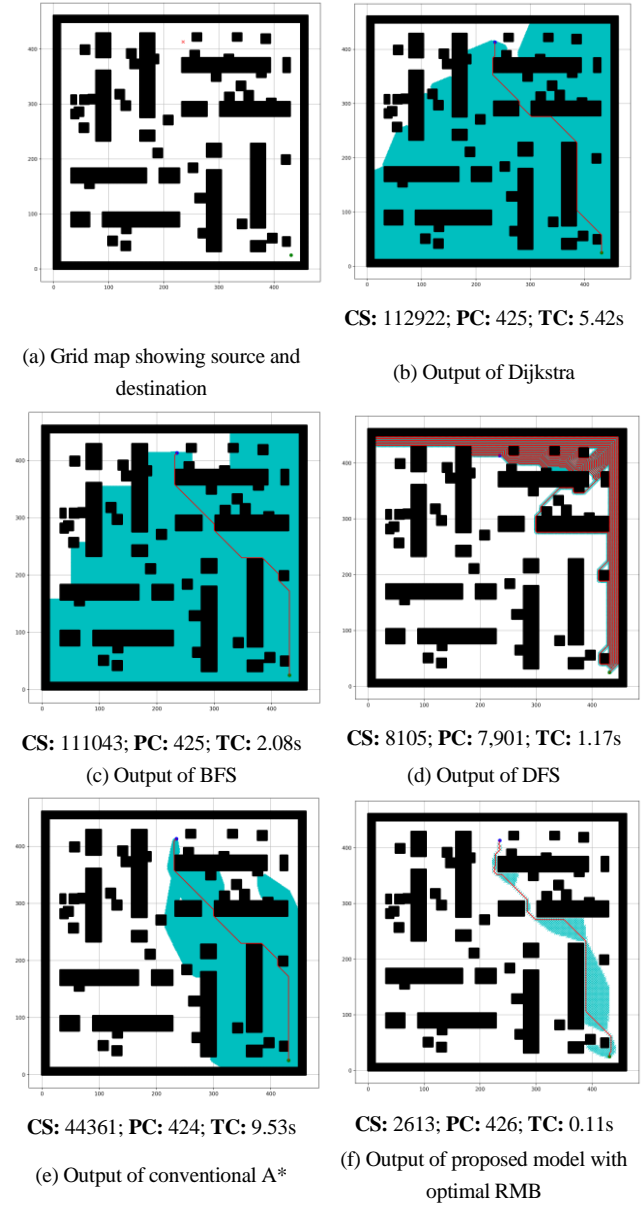

**CS:** 112922; **PC:** 425; **TC:** 5.42s

(a) Grid map showing source and destination

(b) Output of Dijkstra

**CS:** 111043; **PC:** 425; **TC:** 2.08s

(c) Output of BFS

**CS:** 8105; **PC:** 7,901; **TC:** 1.17s

(d) Output of DFS

**CS:** 44361; **PC:** 424; **TC:** 9.53s

(e) Output of conventional A*

**CS:** 2613; **PC:** 426; **TC:** 0.11s

(f) Output of proposed model with optimal RMB

**FIGURE 13.** Simulated results for the comparison between different state-of-the-art algorithms and the proposed model with optimal RMB.

Table IV represents a comprehensive comparison between the proposed model and the different algorithms, Dijkstra, DFS, BFS, and conventional A*. The values from this comparison are the cumulative results collected from the experiment conducted on the considered dataset; thus, these values have fraction numbers. In terms of the number of search cells and time complexity, the proposed model outperformed the other algorithms. Specifically, first, the number of search cells for the proposed model is 2824.45, whereas that for the other algorithms ranges from 46878.88 to 84677.83. Second, regarding time complexity, the proposed model requires 0.1406 seconds, whereas the other algorithms require 1.6375 to 16.5898 seconds. Finally, regarding the path cost, DFS has a value of 4115.53, which is significantly higher than that of the others, and the path cost of the proposed model is mostly similar to that of the remaining algorithms.

TABLE IV
EXPERIMENTAL RESULTS FOR THE COMPARISON BETWEEN DIFFERENT STATE-OF-THE-ART ALGORITHMS AND THE PROPOSED MODEL WITH OPTIMAL RMB.

| Algorithms | Map dimension | Number of search cells | Path cost | Time required (s) |
|---|---|---|---|---|
| **Dijkstra** | **261 x 261** | 43440.25 | 258.95 | 1.5023 |
| | **462 x 261** | 73496.51 | 436.65 | 2.4945 |
| | **462 x 462** | 137696.72 | 477.27 | 7.3836 |
| | *Average* | **84877.83** | **390.96** | **3.7935** |
| **DFS** | **261 x 261** | 36553.30 | 1342.70 | 4.5641 |
| | **462 x 261** | 48019.90 | 5644.20 | 9.5305 |
| | **462 x 462** | 77,399.00 | 5363.70 | 35.6750 |
| | *Average* | **53990.73** | **4116.87** | **16.5898** |
| **BFS** | **261 x 261** | 43874.00 | 258.40 | 1.0516 |
| | **462 x 261** | 72871.50 | 436.65 | 1.2953 |
| | **462 x 462** | 136987.80 | 477.70 | 2.5602 |
| | *Average* | **84577.77** | **390.92** | **1.6357** |
| **A*** | **261 x 261** | 22587.45 | 257.95 | 4.5156 |
| | **462 x 261** | 49740.21 | 435.65 | 12.8063 |
| | **462 x 462** | 68308.97 | 476.70 | 22.3117 |
| | *Average* | **46878.88** | **390.10** | **13.2112** |
| **Proposed model** | **261 x 261** | 2219.61 | 264.85 | 0.0957 |
| | **462 x 261** | 3258.22 | 444.35 | 0.1664 |
| | **462 x 462** | 2995.51 | 470.80 | 0.1597 |
| | *Average* | **2824.45** | **393.33** | **0.1406** |

Figure 14 shows the performance comparison graph of the different algorithms and the proposed model considering the data in Table IV and Equation (9).

$$PC = \frac{\overline{Av}}{Mx} \times 100, \qquad (9)$$

Here, $PC$ = Performance calculation

$\overline{Av}$ = mean of $Av_j$ from $j = 1$ to 3 (map dimensions).

$Mx$ = Maximum value of each parameter.

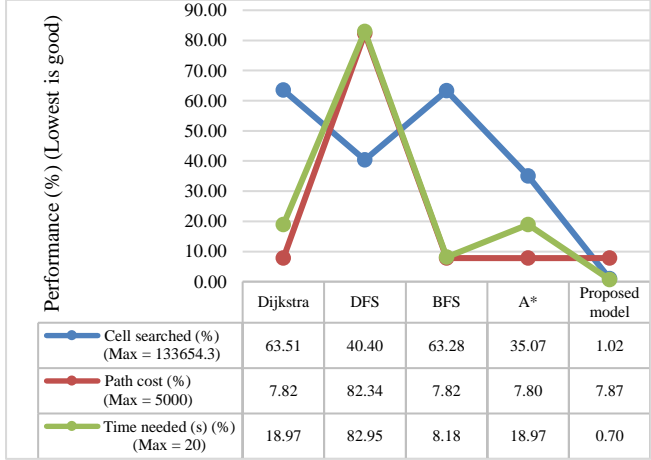

| | Dijkstra | DFS | BFS | A* | Proposed model |
|---|---|---|---|---|---|
| Cell searched (%) (Max = 133654.3) | 63.51 | 40.40 | 63.28 | 35.07 | 1.02 |
| Path cost (%) (Max = 5000) | 7.82 | 82.34 | 7.82 | 7.80 | 7.87 |
| Time needed (s) (%) (Max = 20) | 18.97 | 82.95 | 8.18 | 18.97 | 0.70 |

**FIGURE 14.** Performance of different algorithms compared with the proposed model with optimal RMB.

In the performance calculation, the considered range for the number of search cells is $0 - 133654.33$, as the maximum number of cells in the dataset is 68121 for the map dimension 261 x 261, 120321 for 462 x 261, and 212521 for 462 x 462. The average of these three values is 133654.33. Similarly, The considered ranges for the path cost and time complexity are assumed as follows: $0 - 4500$ and $0 - 20$ seconds, respectively. In this graph, the lowest performance percentage indicates good performance as the lowest values represent the lowest number of search cells, less amount of path cost, and less time complexity to find the target. This table shows that the proposed model outperforms the other algorithms in terms of the number of search cells and time complexity. Its path cost is mostly similar to that of the other algorithms except for DFS, which has the highest path cost.

## C. DISCUSSION

This research aims to enhance the efficiency of the A* path-planning algorithm by minimizing the time complexity and the number of search nodes while maintaining a minimum path cost. The experimental results presented in Fig. 7 and Table I show improved performance compared to the conventional A* algorithm because of its adaptive cost function for the robot movement, which keeps track of the goal position. The reduction in time complexity, the number of search nodes, and path cost is around 29.05 %, 32.58 %, and 0.04 %. According to Table II and Fig. 9, 10, and 11, increasing the size of the proposed RMB reduces the time complexity and the number of search nodes significantly. Fig. 8 samples the simulation results. The path costs increased slightly for the increased size of RMB; however, the time complexity and the number of search nodes decreased. Thus, the proposed model selects the optimal RMB to apply in different applications. The experimental results for the optimal RMB selection are presented in Table III and Fig. 12. According to the experimental results, the RMB $n = 3$ is optimal as it provides a balanced improved results in most cases, as shown in Fig. 12. An additional experiment was conducted, where the proposed model was compared with other similar algorithms, as shown in Table IV, Fig. 13 and 14. The experimental results show that the proposed model outperforms the comparison algorithms. The conventional A* algorithm outperformed the other algorithms. Whereas the performance of the proposed model with optimal RMB ($n = 3$) improved by decreasing the number of search cells 93.98 % and the time complexity 98.94 % compared to the conventional A* algorithm, calculated using Equation (8). The path cost of the proposed model was mostly similar to that of the other algorithms except for DFS, which had the highest path cost.

## V. CONCLUSION

Path planning is crucial for autonomous robot navigation, allowing robots to find optimal routes while avoiding obstacles. A* is a popular global path-planning algorithm due to its heuristic approach. However, conventional A* has limitations, such as computational complexity and extensive cell searching in the grid spaces. This can be problematic for autonomous robots operating in real-time and dynamic environments. To address these limitations, this study proposes an optimal RMB for the A* algorithm that significantly improves its overall performance. In this approach, robot movement costs are calculated by the proposed adaptive cost function that keeps track of the goal to reduce the number of searches to find the goal. Also, the robot checks a certain distance surrounding eight directions of it each time, puts only the end node in the $OPEN$ list of the A* algorithm, and applies a heuristic approach to select the next movement from the $OPEN$ list. These two approaches significantly reduce the number of search cells and time complexity while achieving an acceptable path cost. To evaluate the proposed approach, a benchmarked public dataset with five types of grid maps is used, containing nearly thousands of maps for each type. To generalize the outcome, three different dimensional maps are prepared using the dataset. First, the comparison with the conventional A* shows the improvement of the proposed model. Next, the experimental results show that increasing the proposed RMB size further reduces time complexity and the number of search nodes. Also, the results from the proposed optimal RMB selection method show that the RMB size $n = 3$ is the optimal one, considering the number of search cells, time complexity, and path cost. Finally, the comparison with the different state-of-the-art algorithms shows the effectiveness of the proposed model in improving the performance of the A* algorithm for autonomous robot navigation.

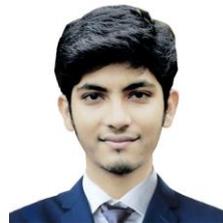

**Raihan Kabir** received his B.Sc. degree in computer science and engineering from the University of Asia Pacific (UAP), Dhaka, Bangladesh, in 2019 and his M.Sc. degree in Computer and Information Systems from the University of Aizu, Fukushima, Japan, in 2021. Currently, he is pursuing his Ph.D. degree in computer science and engineering at the University of Aizu, Fukushima, Japan. His research areas include Machine learning, Image processing, Cloud robotics, and Robotic navigation. He also participated in the World Robot Sumit (WRS) 2020 robotic contest, and his team REL-UoA-JAEA secured 3rd position in the Disaster Robotics Category.

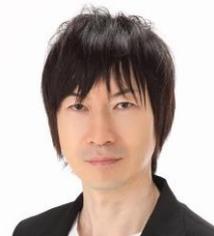

**Yutaka Watanobe** is currently a Senior Associate Professor at the School of Computer Science and Engineering, The University of Aizu, Japan. He received his M.S. and Ph.D. degrees from The University of Aizu in 2004 and 2007, respectively. He was a Research Fellow of the Japan Society for the Promotion of Science (JSPS) at The University of Aizu in 2007. He is now a director of i-SOMET. He was a coach of four ICPC World Final teams. He is a developer of the Aizu Online Judge (AOJ) system. His research interests include intelligent software, programming environments, smart learning, machine learning, data mining, cloud robotics, and visual languages. He is a member of IEEE, IPSJ.

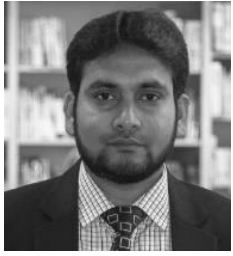

**MD RASHEDUL ISLAM** (Senior Member, IEEE) received the B.Sc. degree in computer science and engineering from the University Rajshahi, Rajshahi, Bangladesh, in 2006, the M.Sc. degree in informatics from the Högskolan i Borås (University of Boras), Boras, Sweden, in 2011, and the Ph.D. degree in electrical, electronic, and computer engineering from the University of Ulsan, Ulsan, South Korea, in 2016. He is currently working as a Chief Researcher of Computer Vision and AI, at Chowagiken Corp., Japan, and also an Associate Professor (on leave) with the Department of Computer Science and Engineering, University of Asia Pacific (UAP), Dhaka, Bangladesh. Previously, he worked as a Senior Architect with the Research and Development Department, Exvision Corporation, Tokyo, Japan, Visiting Researcher (Postdoctoral Researcher) with the School of Computer Science and Engineering, The University of Aizu, Japan; a Graduate Research Assistant with the Embedded System Laboratory, University of Ulsan, South Korea; an Assistant Professor with the Department of Computer Science and Engineering, University of Asia Pacific (UAP), Dhaka, Bangladesh; and a Lecturer with the Department of Computer Science and Engineering, Leading University, Sylhet, Bangladesh. His research interests include machine learning, signal & image processing, HCI, health informatics, bearing fault diagnosis, and others. He is also a PC member of several international conferences. Also, he has good experience in professional IT system analysis and development. He is a member of the IEEE Computer Society and the IEEE Computational Intelligence Society. He has also served as the Secretary of the Organizing Committee of the 19th International Conference on Computer and Information Technology 2017 (ICCIT2017), an Organizing Chair of the Organizing Committee of the ACM-ICPC Dhaka Regional Site 2017, the Head of the Self-Assessment Committee (SAC) of the Department of CSE under IQAC, University of Asia Pacific, a Co-ordinator of the MCSE Program, Department of CSE, University of Asia Pacific, a Convener of Software and Hardware Club, Department of CSE, University of Asia Pacific, a Co-ordinator of the Admission Committee, Department of CSE, University of Asia Pacific, and a Treasurer of the Bangladesh Advanced Computing Society. He is a Reviewer of several journals, such as the IEEE Transactions on Industrial Electronics, IEEE Access, Applied Science, Multimedia Tools and Applications, Cluster Computing, Shock and Vibration, Journal of Information Processing Systems, and others.

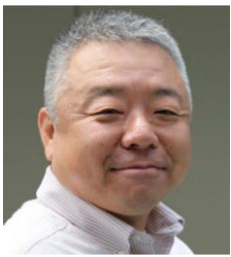

**KEITARO NARUSE** is currently a Full Professor at the School of Computer Science and Engineering at the University of Aizu, Japan. He has worked on the design, development, and standardization of networked distributed intelligent robot systems with heterogeneous robots and sensors. He has applied it to service robot systems, factory automation systems, and intelligent disaster response robot system, which is tested at the Fukushima Robot Test Field. His research interests include swarm robots and their application to agricultural robotic systems and interface systems for disaster response robots.